\documentclass{amia}
\usepackage[T1]{fontenc}
\usepackage{bbm}
\usepackage{multirow}
\usepackage{graphicx,verbatim}
\usepackage{amsmath,amssymb}
\usepackage{booktabs}
\usepackage[hidelinks]{hyperref}
\usepackage{enumitem}
\usepackage{xcolor}

\begin{document}

\title{SDE-Driven Spatio-Temporal Hypergraph Neural Networks for Irregular Longitudinal fMRI Connectome Modeling in Alzheimer's Disease}

\renewcommand{\thefootnote}{\fnsymbol{footnote}}

\author{
Ruiying Chen$^{1,}$\protect\footnotemark[1], 
Yutong Wang$^{2,}$\protect\footnotemark[1], 
Houliang Zhou$^{1}$, 
Wei Liang$^{1}$, 
Yong Chen, PhD$^{3,}$\protect\footnotemark[2], 
Lifang He, PhD$^{1,}$\protect\footnotemark[2]
}

\institutes{
$^{1}$Lehigh University, Bethlehem, PA, USA\\
$^{2}$New York University, New York, NY, USA\\ 
$^{3}$University of Pennsylvania, Philadelphia, PA, USA
}
\maketitle

\footnotetext[1]{Equal contribution.}
\footnotetext[2]{Corresponding authors.}

\renewcommand{\thefootnote}{\arabic{footnote}}
\setcounter{footnote}{0}

\section*{Abstract}
\textit{Longitudinal neuroimaging is essential for modeling disease progression in Alzheimer's disease (AD), yet irregular sampling and missing visits pose substantial challenges for learning reliable temporal representations. To address this challenge, we propose SDE-HGNN, a stochastic differential equation (SDE)-driven spatio-temporal hypergraph neural network for irregular longitudinal fMRI connectome modeling. The framework first employs an SDE-based reconstruction module to recover continuous latent trajectories from irregular observations. Based on these reconstructed representations, dynamic hypergraphs are constructed to capture higher-order interactions among brain regions over time. To further model temporal evolution, hypergraph convolution parameters evolve through SDE-controlled recurrent dynamics conditioned on inter-visit intervals, enabling disease-stage-adaptive connectivity modeling. We also incorporate a sparsity-based importance learning mechanism to identify salient brain regions and discriminative connectivity patterns. Extensive experiments on the OASIS-3 and ADNI cohorts demonstrate consistent improvements over state-of-the-art graph and hypergraph baselines in AD progression prediction. The source code is available at \url{https://github.com/ElektraChen/SDE-HGNN}.}

\section*{Introduction}
\label{sec_intro}
Alzheimer's disease (AD) is a progressive neurodegenerative disorder characterized by gradual cognitive decline and widespread disruption of brain networks. Because neuropathological changes unfold over many years before clinical diagnosis, accurately modeling disease progression is essential for early detection and prognosis prediction~\cite{jack2010hypothetical}. Longitudinal neuroimaging provides repeated observations of brain function over time and offers unique opportunities to characterize subject-specific disease trajectories and heterogeneous progression patterns.

Resting-state functional magnetic resonance imaging (rs-fMRI) enables the study of functional connectivity among distributed brain regions and has been widely used to investigate AD-related network alterations~\cite{smith2004overview}. However, longitudinal neuroimaging data collected in clinical cohorts are inherently irregular, with uneven visit times, variable follow-up intervals, and missing observations. Importantly, such irregularity arises at two temporal scales: (1) \textit{within-scan level}, where ROI-level fMRI time series may be noisy, incomplete, or corrupted due to motion artifacts or signal instability, affecting functional connectivity estimation; and (2) \textit{across-visit level}, where imaging visits occur at irregular intervals or may be missing entirely, making discrete-time disease progression modeling difficult. These characteristics challenge conventional machine learning methods in learning reliable and biologically meaningful temporal representations from irregular longitudinal data.

Graph neural networks (GNNs) have shown promise for brain connectome analysis by representing ROIs as nodes and functional connectivity as edges~\cite{bessadok2022graph,tong2023fmri,dong2023beyond}. However, most existing GNN-based models rely on pairwise graph representations and discrete-time architectures, making them less suitable for irregular longitudinal data. Recent ODE- or SDE-based methods improve continuous-time modeling under irregular sampling~\cite{han2024brainode,zhou2025uncovering}, but they still mainly rely on pairwise graph structures. This limits their ability to capture coordinated activity among multiple brain regions, which is important for modeling higher-order functional organization in AD.

Hypergraphs extend conventional graphs by allowing one hyperedge to connect multiple nodes, providing a natural way to model coordinated multi-region interactions. Existing hypergraph neural networks for brain connectome analysis, such as DwHGCN~\cite{wang2023dynamic}, HyperGALE~\cite{arora2024hypergale}, and HGST~\cite{han2026hypergraph}, mainly focus on static spatial connectivity. Longitudinal hypergraph methods such as WHGCN~\cite{hao2024hypergraph} construct hypergraphs at each time point and aggregate them across predefined visits, but they still rely on discrete-time fusion and model subject-level rather than ROI-level relationships. Thus, existing methods do not jointly model continuous-time disease progression and higher-order ROI interactions in irregular longitudinal fMRI connectomes.

To address these limitations, we propose SDE-HGNN, an SDE-driven spatio-temporal hypergraph neural network for irregular longitudinal fMRI connectome modeling. The proposed framework integrates continuous-time modeling with higher-order brain network representation to capture both temporal disease dynamics and coordinated multi-region functional interactions. Specifically, we first employ an SDE-based module to reconstruct ROI-level fMRI signals as continuous stochastic trajectories, mitigating the effects of irregular or incomplete observations. Based on the reconstructed representations, we construct dynamic hypergraphs that model higher-order relationships among multiple brain regions. To model longitudinal disease progression, hypergraph convolution parameters are further evolved through SDE-controlled recurrent dynamics conditioned on inter-visit intervals. In addition, we incorporate a sparsity-based importance learning mechanism to identify salient brain regions and discriminative connectivity patterns associated with AD progression. SDE-HGNN is the first framework that jointly models continuous-time longitudinal dynamics and higher-order brain network interactions for irregular fMRI connectome analysis.

The main contributions of this work are summarized as follows: 
\vspace{-10pt}
\begin{itemize}[itemsep=0pt]
    \item We introduce an SDE-based reconstruction module to model ROI-level fMRI signals as continuous stochastic trajectories, addressing irregular or incomplete observations in fMRI time series. 
    \item We develop a novel SDE-guided spatio-temporal hypergraph neural network in which hypergraph convolution parameters evolve continuously according to inter-visit intervals, enabling adaptive modeling of disease progression under irregular longitudinal sampling.
    \item We incorporate a sparsity-based learning mechanism within the hypergraph framework to identify salient brain regions and disease-specific connectivity patterns, providing interpretable insights into AD progression. 
    \item Extensive experiments on the OASIS-3 and ADNI cohorts demonstrate consistent improvements over state-of-the-art graph and hypergraph baselines while revealing clinically meaningful longitudinal connectivity biomarkers aligned with known AD pathology.
\end{itemize}

\section*{Methodology}
\label{sec_method}
\begin{figure*}[t]
    \centering
    \includegraphics[width=\textwidth]{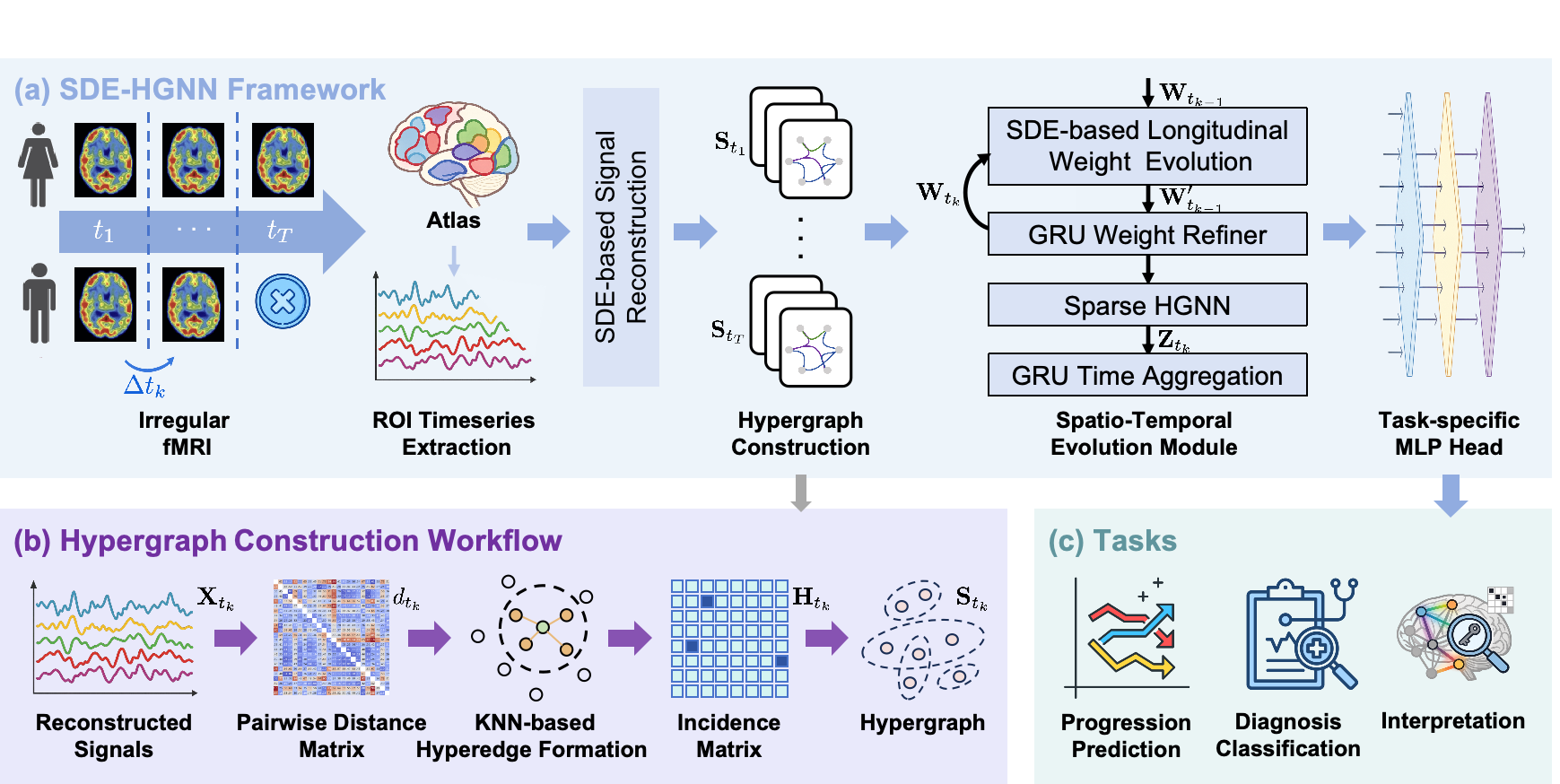}
    \caption{Overview of SDE-HGNN framework. (a) Irregular longitudinal fMRI signals are reconstructed using a neural SDE, followed by hypergraph construction and SDE-driven spatio-temporal modeling with GRU refinement and aggregation. (b) Hypergraph construction from reconstructed ROI signals via pairwise distance computation and KNN-based hyperedge formation. (c) Downstream tasks include progression prediction, diagnosis and interpretation.}
    \label{fig:framework}
\end{figure*}

We consider the problem of predicting disease progression from irregular longitudinal rs-fMRI connectomes. Suppose that each subject contains multiple imaging visits acquired at irregular time intervals. For a subject with up to $T$ visits occurring at time points $\{t_1, t_2, \cdots, t_T\}$, the inter-visit interval is defined as $\Delta t_k = t_k - t_{k-1}$. For each visit, ROI-level BOLD time series are extracted after standard preprocessing and brain parcellation. Let $N$ denote the number of regions of interest (ROIs) and $D$ the feature dimension per ROI. After preprocessing, the ROI feature matrix at visit $t_k$ is denoted by $\mathbf{X}_{t_k} \in \mathbb{R}^{N \times D}$, where each row corresponds to the feature representation of one brain region. As discussed in the Introduction, irregularity in longitudinal rs-fMRI arises at two temporal scales: noisy or incomplete ROI time series within scans and heterogeneous intervals between visits. To address these challenges, our proposed SDE-HGNN framework consists of three sequential stages: (1) SDE-based fMRI signal reconstruction, (2) Hypergraph construction at each visit, and (3) SDE-guided spatio-temporal hypergraph modeling. An overview of the SDE-HGNN framework is shown in Fig.~\ref{fig:framework}. 

Intuitively, the first stage transforms irregular and noisy ROI-level fMRI signals into temporally coherent latent trajectories, so that functional representations can be reconstructed and compared across visits. In the second stage, a visit-specific hypergraph is constructed from the reconstructed ROI representations to capture coordinated interactions among multiple brain regions, going beyond conventional pairwise connectivity modeling. In the third stage, the hypergraph convolution weights are dynamically evolved according to inter-visit intervals, allowing the model to adapt its connectivity representation to subject-specific disease progression rates under irregular longitudinal sampling. We describe each stage in detail below.

\textbf{\textit{SDE-based fMRI Signal Reconstruction.}}
\label{sec_recon}
Resting-state fMRI signals are inherently noisy and may contain irregular or incomplete observations. To obtain temporally coherent ROI-level representations, we model each ROI's BOLD time series as a continuous stochastic process. Specifically, we adopt a neural stochastic differential equation (SDE) encoder–decoder framework~\cite{de2019gru}. A recurrent neural network (RNN) encoder first processes the irregularly sampled observations and maps them to an approximate posterior over an initial latent state $\mathbf{z}_0$. A neural SDE then models the latent trajectory as
\begin{equation}
d\mathbf{z}_t = f_{\theta}(\mathbf{z}_t, t)\, dt + g_{\theta}(\mathbf{z}_t, t)\, d{B}_t
\end{equation}
where $t$ denotes continuous time within a scan, $f_{\theta}$ is a neural network–parameterized drift function modeling deterministic temporal dynamics, $g_{\theta}$ is a learnable diffusion function controlling stochastic variability in the latent process, and ${B}_t$ denotes standard Brownian motion. 

The learned SDE enables interpolation at arbitrary time points, generating continuous latent trajectories $\mathbf{z}_t$ even under irregular sampling. A decoder then reconstructs complete ROI-level time series from these latent states. This reconstruction step mitigates artifacts caused by irregular intra-scan sampling and ensures that downstream functional connectivity estimation is based on temporally consistent signals. After reconstruction, each visit $t_k$ is represented by an ROI feature matrix $\mathbf{X}_{t_k} \in \mathbb{R}^{N \times D}$.

\textbf{\textit{Hypergraph Construction.}}
\label{sec_hypergraph}
Standard connectome modeling constructs pairwise connectivity graphs via ROI-\allowbreak ROI correlations. However, many neural processes arise from coordinated activity among multiple brain regions simultaneously. To explicitly model higher-order interactions, we construct a hypergraph at each visit, which generalizes the standard graph by allowing each hyperedge to connect an arbitrary subset of nodes. For each hypernode $v_i$, we compute pairwise Euclidean distances between node features and select its $K$-nearest neighbors (KNN) to form a hyperedge $e_j$. Let $\mathbf{x}_{t_k}^{(i)} \in \mathbb{R}^D$ denote the feature vector of ROI $i$, corresponding to the $i$-th row of $\mathbf{X}_{t_k}$. The hypergraph incidence matrix at visit $t_k$ is defined as
\begin{equation}
\label{eq_H_entry}
H_{t_k}(i, j) =
\begin{cases}
\exp\!\bigl(-\frac{{d_{t_k}^{(ij)}}^2}{(q \cdot \bar{d}_{t_k}^{(i)})^{2}}\bigr), & v_i \in \mathcal{N}_K(v_i) \\[2pt]
0, & \text{otherwise}
\end{cases}
\end{equation}
where $d_{t_k}^{(ij)} = 
\|\mathbf{x}_{t_k}^{(i)} - \mathbf{x}_{t_k}^{(j)}\|_2$ and $\bar{d}_{t_k}^{(i)}$ denotes the mean distance from hypernode $v_i$ to all other hypernodes, $q$ is a 
scaling parameter, and $\mathcal{N}_K(v_i)$ denotes the $K$-nearest-neighbor set of hypernode $v_i$. Each hypernode participates in multiple overlapping hyperedges, enabling message passing beyond pairwise relationships.

Let $\mathbf{M}_e = \operatorname{diag}(m_1, \ldots, m_E) \in \mathbb{R}^{E \times E}$ denote the diagonal hyperedge weight matrix, where $m_j$ is the weight of hyperedge $e_j$ (initialized to 1 and later modulated by learned importance probabilities), and $E$ denotes the number of hyperedges. Hypernode degrees and hyperedge degrees are defined as
\begin{equation}
    d(v_i) = \sum_{j=1}^{E} m_j\, H_{t_k}(i, j), \quad \delta(e_j) = \sum_{i=1}^{N} H_{t_k}(i, j).
\end{equation}
Let $\mathbf{D}_v \in \mathbb{R}^{N \times N}$ and $\mathbf{D}_e \in \mathbb{R}^{E \times E}$ denote the diagonal hypernode and hyperedge degree matrices. The normalized hypergraph propagation matrix can be expressed as
\begin{equation}
\label{eq:S}
\mathbf{S}_{t_k} = \mathbf{D}_v^{-1/2}\, \mathbf{H}_{t_k}\, \mathbf{M}_e\, \mathbf{D}_e^{-1}\, \mathbf{H}_{t_k}^{\mathrm{T}}\, \mathbf{D}_v^{-1/2}.
\end{equation}
The propagation matrix $\mathbf{S}_{t_k}$ encodes higher-order neighborhood relationships for downstream hypergraph convolution.

\label{sec_model}
\textbf{\textit{SDE-guided Spatio-temporal HGNN.}} To model longitudinal disease progression under irregular visit intervals, we allow the hypergraph convolution parameters to evolve continuously over time. Let $\mathbf{W}_{t_k} \in \mathbb{R}^{D \times d_l}$ denote the hypergraph convolution weight state at visit $t_k$, where $d_l$ is the number of output dimensions. We model its evolution as an SDE-driven continuous-time stochastic process
\begin{equation}
    \mathbf{W}'_{t_k} = \operatorname{SDESolve}(f_{\theta}, \mathbf{W}_{t_{k-1}}, \Delta t_k),
\end{equation}
where $f_{\theta}$ is a learnable drift function and $\Delta t_k$ is the inter-visit time interval.

To incorporate visit-specific information, we further refine the evolved weights using a gated recurrent unit (GRU) module conditioned on the current ROI features:
\begin{equation}
\mathbf{W}_{t_k} = \operatorname{GRU}(\mathbf{X}_{t_k}, \mathbf{W}'_{t_k}) + \operatorname{GRU}(\mathbf{X}_{t_k}, \mathbf{W}_0),
\end{equation}
where the second term introduces a time-invariant baseline initialized by a learnable weight state $\mathbf{W}_0$ shared across visits, allowing the convolution parameters to adapt dynamically to heterogeneous disease progression rates.

The refined weight matrix $\mathbf{W}_{t_k}$ is then used in the hypergraph convolution operator to compute visit-specific hypernode representations:
\begin{equation}
\label{eq:sde_evolve}
\mathbf{Z}_{t_k} = \sigma\!\bigl(\mathbf{S}_{t_k}\mathbf{X}_{t_k}\mathbf{W}_{t_k} \bigr),
\end{equation}
where $\mathbf{Z}_{t_k} \in \mathbb{R}^{N \times d_l}$ is the output hypernode embeddings, and $\sigma$ is the activation function. Through this process, the model captures higher-order interactions among brain regions while jointly modeling continuous-time disease progression in irregular longitudinal neuroimaging data.

\textbf{\textit{Sparse Interpretability.}} The hypergraph structure $\mathbf{S}_{t_k}$ and hypernode feature matrix $\mathbf{X}_{t_k}$ may contain redundant or noisy information that is not essential for disease prediction. To identify informative brain regions and connectivity patterns, we introduce a sparsity-based interpretability mechanism that learns importance scores for both hypernode features and hyperedges. Specifically, inspired by GNNExplainer~\cite{ying2019gnnexplainer}, we learn the hypernode feature importance probabilities $\mathbf{P}_X \in \mathbb{R}^{N \times D}$ and hyperedge importance probabilities $\mathbf{P}_E \in \mathbb{R}^{E}$. The sparsified hypernode features are obtained by
\begin{equation}
    \tilde{\mathbf{X}}_{t_k} = \mathbf{X}_{t_k} \odot \mathbf{P}_X,
\end{equation}
where $\odot$ denotes the element-wise multiplication. 

Let $\mathbf{P}_E = [p_{E_1}, p_{E_2}, \ldots, p_{E_N}]$ denote the vector of hyperedge importance probabilities, where $p_{E_i}$ represents the importance score of hyperedge $e_i$. Each hyperedge importance probability is computed from the importance-weighted feature of its center node:
\begin{equation}
\label{eq:hedge_prob}
p_{E_i} = \sigma\bigl(\mathbf{v}^{\mathrm{T}} (\mathbf{x}_{t_k}^{(i)} \odot \mathbf{p}_i)\bigr),
\end{equation}
where $\mathbf{v} \in \mathbb{R}^{D}$ is a learnable projection vector, and $\mathbf{p}_i \in \mathbb{R}^D$ is the $i$-th row of the hypernode feature importance matrix $\mathbf{P}_X$, corresponding to the feature importance weights of hypernode $i$.

Using these learnable importance probabilities, the hypergraph propagation matrix is recomputed as
\begin{equation}
\label{eq:S_sparse}
\tilde{\mathbf{S}}_{t_k} = \tilde{\mathbf{D}}_v^{-1/2} \mathbf{H}_{t_k}\operatorname{diag}({p}_{E_1}, \dots, {p}_{E_N}) \tilde{\mathbf{D}}_e^{-1}\, \mathbf{H}_{t_k}^{\mathrm{T}} \tilde{\mathbf{D}}_v^{-1/2},
\end{equation}
Finally, the sparsified hypernode representations are computed as
\begin{equation}
\label{eq:sparse_forward}
\hat{\mathbf{Z}}_{t_k} = \sigma\!\bigl(\tilde{\mathbf{S}}_{t_k}\tilde{\mathbf{X}}_{t_k}\mathbf{W}_{t_k}\bigr).
\end{equation}
which replaces the original convolution in Eq.~(\ref{eq:sde_evolve}). This sparsification mechanism enables the model to identify salient brain regions and discriminative connectivity patterns associated with disease progression.

\textbf{\textit{Loss Function.}} After hypergraph convolution, node embeddings $\hat{\mathbf{Z}}_{t_k}$ are aggregated via global max and mean pooling to obtain graph-level representations. The resulting features are fed to a multilayer perceptron (MLP) classifier for disease progression prediction. The overall training objective combines classification loss with sparsity and information-preserving regularization:
\begin{equation}
\label{eq:loss}
\mathcal{L} = \mathcal{L}_{ce} + \lambda_1 \mathcal{L}_{mi} + \lambda_2 \mathcal{L}_s + \lambda_3 \mathcal{L}_e.
\end{equation}
The classification loss is defined as
\begin{equation}
    \mathcal{L}_{ce} = -\bigl(y \log \sigma(\hat{y}) + (1 - y) \log (1 - \sigma(\hat{y})\bigr).
\end{equation}
To ensure that the sparsified hypernode features and hypergraph structure preserve sufficient predictive information, we introduce a mutual information regularization term
\begin{equation}
\label{eq:loss_mi}
\mathcal{L}_{mi} = -\sum_{c=1}^{C} \mathbbm{1}[y=c] \log{P}_{\Phi}\!\bigl(\hat{y}={y} \mid \tilde{\mathbf{S}}_{t_k},\; \tilde{\mathbf{X}}_{t_k}\bigr),
\end{equation}
which maximizes the mutual information between the sparsified representations and the disease labels. 

The sparsity regularization is defined as
\begin{equation}
\label{eq:loss_sp}
\mathcal{L}_s = \|\mathbf{P}_X\|_1 + \frac{1}{T}\sum_{k=1}^{T}\|\mathbf{P}_{E}\|_1.
\end{equation}
To encourage discrete feature selection, we further apply binary entropy regularization: $\mathcal{L}_e = \mathcal{L}_{P_X} + \mathcal{L}_{P_E}$, where
\begin{align}
    &\mathcal{L}_{P_X} = -\bigl(\mathbf{P}_X \log(\mathbf{P}_X) + (1 - \mathbf{P}_X) \log(1 - \mathbf{P}_X)\bigr),
    & \mathcal{L}_{P_E} = -\frac{1}{T}\sum_{k=1}^{T}\bigl(\mathbf{P}_E \log(\mathbf{P}_E) + (1 - \mathbf{P}_E) \log(1 - \mathbf{P}_E)\bigr).
\end{align}
Together, these losses encourage the model to identify a compact subset of salient ROIs and hyperedges that are most informative for predicting disease progression. The coefficients $\lambda_1$, $\lambda_2$, and $\lambda_3$ are tunable hyperparameters that control the trade-off among the different loss terms.

\begin{table}[t]
\centering
\caption{Classification comparisons for stable vs. progressive subjects on OASIS-3.}
\label{tab:oasis_points}
\renewcommand{\arraystretch}{1.2}
\resizebox{\textwidth}{!}{
\begin{tabular}{ccccccc}
\toprule[1.5pt] 
& Num. of Timepoints & Methods & AUC & Accuracy & Sensitivity & Specificity \\
\midrule 
& \multirow{4}{*}{Cross-sectional}
& DGCNN & $0.6812 \pm 0.0094$ & $0.6500 \pm 0.0240$ & $0.6246 \pm 0.0946$ & $0.6587 \pm 0.0629$ \\
& & DwHGCN & $0.6168 \pm 0.0346$ & $0.6988 \pm 0.0095$ & $0.3275 \pm 0.1117$ & $0.8286 \pm 0.0428$ \\
& & HyperGALE & $0.6408 \pm 0.0483$ & $0.6614 \pm 0.0150$ & $ 0.4365 \pm 0.0094$ & $0.7250 \pm 0.0280$ \\ 
& & HGST & $0.6348 \pm 0.0516$ & $0.7162 \pm 0.0230$ & $0.5707 \pm 0.0368$ & $0.8718 \pm 0.0195$ \\
\cmidrule(lr){2-7} 
& \multirow{4}{*}{1} 
& BrainTokenGT & $0.4840 \pm 0.0403$ & $0.5369 \pm 0.2150$ & $0.3853 \pm 0.4239$ & $0.5874 \pm 0.4364$ \\
& & WHGCN & $0.5539 \pm 0.0748$ & $0.5292 \pm 0.1390$ & $\mathbf{0.5278 \pm 0.2461}$ & $0.5292 \pm 0.2613$ \\
& & SDEGCN & $0.5702 \pm 0.0598$ & $0.6706 \pm 0.0706$ & $0.2385 \pm 0.1122$ & $0.8213 \pm 0.1323$ \\
& & \textbf{SDE-HGNN} & $\mathbf{0.6067 \pm 0.0458}$ & $\mathbf{0.7148 \pm 0.0460}$ & $0.2324 \pm 0.1575$ & $\mathbf{0.8825 \pm 0.0905}$ \\
\cmidrule(lr){2-7} 
& \multirow{4}{*}{2} 
& BrainTokenGT & $0.5230 \pm 0.0252$ & $0.5061 \pm 0.0547$ & $0.5480 \pm 0.1240$ & $0.4912 \pm 0.1085$ \\
& & WHGCN & $0.5848 \pm 0.0690$ & $0.5967 \pm 0.1482$ & $0.5457 \pm 0.2451$ & $0.6155 \pm 0.2749$ \\
& & SDEGCN & $0.6738 \pm 0.0396$ & $0.6236 \pm 0.1818$ & $\mathbf{0.5865 \pm 0.2165}$ & $0.6374 \pm 0.3173$ \\
& & \textbf{SDE-HGNN} & $\mathbf{0.7293 \pm 0.0206}$ & $\mathbf{0.7322 \pm 0.0304}$ & $0.5648 \pm 0.0182$ & $\mathbf{0.7906 \pm 0.0374}$ \\
\cmidrule(lr){2-7} 

& \multirow{4}{*}{3} 
& BrainTokenGT & $0.5330 \pm 0.0339$ & $0.5314 \pm 0.0882$ & $0.4663 \pm 0.1734$ & $0.5540 \pm 0.1719$ \\
& & WHGCN & $0.5954 \pm 0.0438$ & $0.5997 \pm 0.0692$ & $0.5958 \pm 0.1046$ & $0.6010 \pm 0.1230$ \\
& & SDEGCN & $0.7103 \pm 0.0271$ & $0.6960 \pm 0.0779$ & $0.5448 \pm 0.1533$ & $\mathbf{0.7493 \pm 0.1539}$ \\
& & \textbf{SDE-HGNN} & $\mathbf{0.7726 \pm 0.0361}$ & $\mathbf{0.7081 \pm 0.0642}$ & $\mathbf{0.6987 \pm 0.0973}$ & $0.7108 \pm 0.1153$ \\
\cmidrule(lr){2-7} 

& \multirow{4}{*}{4} 
& BrainTokenGT & $0.5410 \pm 0.0649$ & $0.5741 \pm 0.0500$ & $0.4861 \pm 0.0942$ & $0.6045 \pm 0.0715$ \\
& & WHGCN & $0.6148 \pm 0.0213$ & $0.6079 \pm 0.0739$ & $0.5803 \pm 0.1033$ & $0.6175 \pm 0.1330$ \\
& & SDEGCN & $0.7305 \pm 0.0311$ & $0.7095 \pm 0.0202$ & $0.5811 \pm 0.0863$ & $0.7545 \pm 0.0181$ \\
& & \textbf{SDE-HGNN} & $\mathbf{0.7769 \pm 0.0337}$ & $\mathbf{0.7496 \pm 0.0409}$ & $\mathbf{0.6629 \pm 0.0423}$ & $\mathbf{0.7795 \pm 0.0695}$ \\
\cmidrule(lr){2-7} 

& \multirow{4}{*}{5} 
& BrainTokenGT & $0.5758 \pm 0.0244$ & $0.5866 \pm 0.0756$ & $0.4709 \pm 0.2040$ & $0.6268 \pm 0.1733$ \\
& & WHGCN & $0.6153 \pm 0.0363$ & $0.6198 \pm 0.0838$ & $0.5753 \pm 0.1385$ & $0.6355 \pm 0.1546$ \\
& & SDEGCN & $0.7153 \pm 0.0234$ & $\mathbf{0.7135 \pm 0.0518}$ & $0.5287 \pm 0.1178$ & $\mathbf{0.7780 \pm 0.1022}$ \\
& & \textbf{SDE-HGNN} & $\mathbf{0.7808 \pm 0.0354}$ & $\mathbf{0.7135 \pm 0.0462}$ & $\mathbf{0.6833 \pm 0.0767}$ & $0.7238 \pm 0.0791$ \\
\cmidrule(lr){2-7} 

& \multirow{4}{*}{6} 
& BrainTokenGT & $0.5489 \pm 0.0416$ & $0.5649 \pm 0.0269$ & $0.4973 \pm 0.0798$ & $0.5885 \pm 0.0147$ \\
& & WHGCN & $0.6312 \pm 0.0363$ & $0.6506 \pm 0.0471$ & $0.5182 \pm 0.1978$ & $0.6965 \pm 0.1216$ \\
& & SDEGCN & $0.7335 \pm 0.0265$ & $0.7148 \pm 0.0393$ & $0.5345 \pm 0.1181$ & $\mathbf{0.7780 \pm 0.0781}$ \\
& & \textbf{SDE-HGNN} & $\mathbf{0.7772 \pm 0.0297}$ & $\mathbf{0.7255 \pm 0.0317}$ & $\mathbf{0.6785 \pm 0.0367}$ & $0.7417 \pm 0.0511$ \\

\bottomrule[1.5pt] 
\end{tabular}
}
\end{table}
\section*{Experiments and Results}
\label{sec_expriments}
We evaluate SDE-HGNN along four aspects: (i) longitudinal disease progression prediction under varying follow-up lengths on the OASIS-3 dataset; (ii) diagnostic classification on the ADNI dataset; (iii) ablation studies to analyze the contributions of key model components; and (iv) saliency visualization to assess model interpretability. 

\textbf{\textit{Datasets.}} We conduct experiments on two publicly available longitudinal neuroimaging datasets: OASIS-3~\cite{marcus2010open} and ADNI~\cite{mueller2005alzheimer}. The OASIS-3 cohort contains 747 subjects, including 193 progressive subjects and 554 stable subjects. Each subject contributes up to six irregular longitudinal visits, which are mapped to cohort-mean follow-up months: baseline, 34, 62, 74, 81, and 93 months. The irregular intervals between visits make this dataset particularly suitable for evaluating continuous-time disease progression models. The ADNI cohort includes 289 subjects (135 healthy controls, 100 MCI, and 54 AD). We evaluate three binary diagnostic tasks: HC vs.\ AD, HC vs.\ MCI, and MCI vs.\ AD.

\textbf{\textit{Experimental Setup.}} For both datasets, rs-fMRI scans were preprocessed using the standard fMRIPrep pipeline~\cite{esteban2019fmriprep}, including slice-timing correction, MCFLIRT-based motion correction, and spatial normalization to the MNI152 template. To further reduce motion artifacts, ICA-AROMA~\cite{pruim2015ica} was applied, followed by spatial smoothing with a 6 mm FWHM Gaussian kernel. The cortex was parcellated using the Schaefer-100 atlas~\cite{schaefer2018local}, yielding 100 regions of interest (ROIs). ROI-level BOLD time series were extracted and used to construct node features for hypergraph modeling. To prevent data leakage and ensure predictive validity, only longitudinal scans prior to the point of clinical conversion were used for progressive subjects. 

Functional connectivity representations were derived from the reconstructed ROI signals produced by the SDE module. Hypergraphs were constructed at each visit using a KNN strategy based on pairwise distances between ROI features. The number of neighbors $K$ was set to 10. The proposed model was implemented in PyTorch and trained using the Adam optimizer. The learning rate, hidden dimensions, and regularization coefficients were selected via validation on the training folds. All experiments were conducted using five-fold cross-validation with identical stratified subject-level splits (6:2:2 ratio), and the reported results correspond to the average performance across folds. Performance is measured using the following metrics: area under the ROC curve (AUC), accuracy, sensitivity, and specificity.

\begin{table}[t]
\centering
\caption{Classification performance for different methods on ADNI dataset.}
\label{tab:ADNI_baselines}
\renewcommand{\arraystretch}{1.2}
\resizebox{\textwidth}{!}{
\begin{tabular}{ccccccc}
\toprule[1.5pt] 
& Tasks & Methods & AUC & Accuracy & Sensitivity & Specificity \\
\midrule

& \multirow{8}{*}{HC vs. AD}
& DGCNN & $0.5817 \pm 0.1661$ & $0.6797 \pm 0.0993$ & $0.3321 \pm 0.2983$ & $0.7733 \pm 0.1111$ \\
& & DwHGCN & $0.6312 \pm 0.0362$ & $0.6195 \pm 0.0372$ & $0.4653 \pm 0.0635$  & $0.7658 \pm 0.0897$ \\
& & BrainTokenGT & $0.5770 \pm 0.1119$ & $0.5932 \pm 0.0844$ & $0.5222 \pm 0.2603$ & $0.6138 \pm 0.1613$ \\
& & WHGCN & $0.6039 \pm 0.1126$ & $0.5453 \pm 0.1298$ & $0.5333 \pm 0.3943$ & $0.5551 \pm 0.3273$ \\
& & HyperGALE & $0.6102 \pm 0.0183$ & $0.6769 \pm 0.0625$ & $0.4779 \pm 0.0724$ & $0.7872 \pm 0.0729$ \\
& & SDEGCN & $0.6504 \pm 0.0435$ & $\mathbf{0.7141 \pm 0.0225}$ & $0.4291 \pm 0.1957$ & $\mathbf{0.8296 \pm 0.0895}$ \\
& & HGST & $0.6036 \pm 0.1163$ & $0.6317 \pm 0.0908$ & $0.3882 \pm 0.1122$ & $0.7576 \pm 0.1515$ \\
& & \textbf{SDE-HGNN} & $\mathbf{0.6989 \pm 0.1008}$ & $0.6770 \pm 0.0530$ & $\mathbf{0.5382 \pm 0.1101}$ & $0.7333 \pm 0.1032$ \\
\cmidrule(lr){2-7}

& \multirow{8}{*}{HC vs. MCI}
& DGCNN & $0.5421 \pm 0.1862$ & $0.6308 \pm 0.1376$ & $0.4071 \pm 0.2405$ & $0.6913 \pm 0.1844$ \\
& & DwHGCN & $0.5876 \pm 0.0087$  & $0.6277 \pm 0.0038$ & $0.4433 \pm 0.0522$ & $0.6335 \pm 0.0528$ \\
& & BrainTokenGT & $0.5628 \pm 0.0654$ & $0.5421 \pm 0.0779$ & $0.5130 \pm 0.0748$ & $0.5664 \pm 0.1267$ \\
& & WHGCN & $0.5674 \pm 0.0563$ & $0.5421 \pm 0.0887$ & $0.4846 \pm 0.3149$ & $0.6375 \pm 0.2860$ \\
& & HyperGALE & $0.5733 \pm 0.0251$ & $0.5953 \pm 0.0152$ & $0.4645 \pm 0.0627$ & $0.6638 \pm 0.0142$ \\
& & SDEGCN & $0.5948 \pm 0.0740$ & $0.5915 \pm 0.0313$ & $0.3000 \pm 0.2588$ & $\mathbf{0.8074 \pm 0.1596}$ \\
& & HGST & $0.5639 \pm 0.1513$ & $0.6072 \pm 0.1008$ & $0.4154 \pm 0.0987$ & $0.6739 \pm 0.2219$ \\
& & \textbf{SDE-HGNN} & $\mathbf{0.6241 \pm 0.0480}$ & $\mathbf{0.6468 \pm 0.0478}$ & $\mathbf{0.5500 \pm 0.1140}$ & $0.7185 \pm 0.0763$ \\
\cmidrule(lr){2-7}

& \multirow{8}{*}{MCI vs. AD}
& DGCNN & $0.5295 \pm 0.1633$ & $0.6414 \pm 0.1012$ & $0.3786 \pm 0.2087$ & $0.7117 \pm 0.1432$ \\
& & DwHGCN & $0.5325 \pm 0.0076$ & $0.6100 \pm 0.0073$ & $0.4142 \pm 0.0575$ & $0.6767 \pm 0.0032$ \\
& & BrainTokenGT & $0.5372 \pm 0.0993$ & $0.5116 \pm 0.0834$ & $\mathbf{0.4964 \pm 0.2447}$ & $0.5193 \pm 0.1782$ \\
& & WHGCN & $0.5595 \pm 0.0795$ & $0.5746 \pm 0.2189$ & $0.3714 \pm 0.3452$ & $0.6231 \pm 0.3628$ \\
& & HyperGALE & $0.5682 \pm 0.0823$ & $0.6174 \pm 0.0526$ & $0.3587 \pm 0.0263$ & $0.7342 \pm 0.0198$ \\
& & SDEGCN & $0.5274 \pm 0.0683$ & $0.5460 \pm 0.1059$ & $0.4018 \pm 0.4537$ & $0.6200 \pm 0.4057$ \\
& & HGST & $0.5349 \pm 0.0948$ & $0.6233 \pm 0.2016$ & $0.3766 \pm 0.2313$ & $0.7019 \pm 0.1522$ \\
& & \textbf{SDE-HGNN} & $\mathbf{0.5873 \pm 0.1414}$ & $\mathbf{0.6495 \pm 0.0086}$ & $0.2455 \pm 0.2909$ & $\mathbf{0.8200 \pm 0.1600}$ \\
\bottomrule[1.5pt]
\end{tabular}
}
\end{table}

\textbf{\textit{Comparison with Competing Methods.}} 
We compare SDE-HGNN with both cross-sectional and longitudinal baselines. Cross-sectional baselines include DGCNN~\cite{zhao2022dynamic}, DwHGCN~\cite{wang2023dynamic}, HyperGALE~\cite{arora2024hypergale}, HGST~\cite{han2026hypergraph}. Longitudinal baselines include Brain-TokenGT~\cite{dong2023beyond}, WHGCN~\cite{hao2024hypergraph}, SDEGCN~\cite{zhou2025uncovering}. These baselines cover a range of graph and hypergraph neural network approaches for brain connectome analysis.

Table~\ref{tab:oasis_points} reports progression prediction results on OASIS-3 under varying numbers of longitudinal visits. SDE-HGNN consistently achieves the best performance across all settings. The performance gap becomes more pronounced as the number of visits increases, demonstrating the effectiveness of SDE-based continuous-time modeling in capturing longitudinal disease dynamics. In addition, the hypergraph formulation enables the model to capture higher-order interactions among multiple brain regions, leading to stronger performance even when only a small number of visits are available. Table~\ref{tab:ADNI_baselines} reports diagnostic classification results on ADNI. While the OASIS-3 experiments primarily evaluate longitudinal modeling capability, the ADNI experiments emphasize spatial representation learning. Across all diagnostic tasks, SDE-HGNN achieves the highest AUC and accuracy. In particular, the model shows clear advantages in the clinically challenging early-stage scenarios (HC vs. MCI and MCI vs. AD), indicating that the hypergraph representation effectively captures subtle multi-region functional abnormalities.

\begin{table}[t]
\centering
\caption{Ablation results for temporal modeling and sparsity modules on the OASIS-3 dataset (progressive vs. stable classification with six time points).}
\label{tab:ablation_temporal}
\setlength{\tabcolsep}{3.5pt}
\resizebox{\columnwidth}{!}{
\begin{tabular}{lcccc}
\toprule
Methods & AUC & Accuracy & Sensitivity & Specificity \\
\midrule
HGNN+RNN & $0.7328\pm0.0370$ & $0.6760\pm0.0461$ & $0.6078\pm0.2949$ & $0.7001\pm0.1526$ \\
HGNN+ODE & $0.7519\pm0.0414$ & $0.7161\pm0.0722$ & $0.6269\pm0.0216$ & $0.7472\pm0.0941$ \\
SDE-HGNN (w/o Sparsity) & $0.7556\pm0.0364$ & $0.7161\pm0.0670$ & $0.6069\pm0.1510$ & $\mathbf{0.7541\pm0.1365}$ \\
\textbf{SDE-HGNN} (Full) & $\mathbf{0.7772\pm0.0297}$ & $\mathbf{0.7255\pm0.0317}$ & $\mathbf{0.6785\pm0.0367}$ & $0.7417\pm0.0511$ \\
\bottomrule
\end{tabular}
}
\end{table}

\textbf{\textit{Ablation Studies.}} To assess the contribution of key components in SDE-HGNN, we conduct ablation studies on OASIS-3 by modifying the temporal evolution strategy and the sparsity mechanism. As shown in Table~\ref{tab:ablation_temporal}, two observations can be made. First, replacing the SDE-driven temporal evolution with a discrete-time RNN or deterministic ODE leads to a noticeable performance drop. This indicates that continuous-time stochastic modeling is important for capturing subject-specific variability in irregular longitudinal trajectories. Second, removing the sparsity regularization terms ($\mathcal{L}_{s}$, $\mathcal{L}_{e}$) also reduces performance. This suggests that the interpretability mechanism not only provides explanations but also acts as an effective regularizer that suppresses redundant connectivity patterns, thereby improving predictive performance.



\begin{figure*}[h!]
    \centering
    \vspace{-5pt}
    \includegraphics[width=\columnwidth]{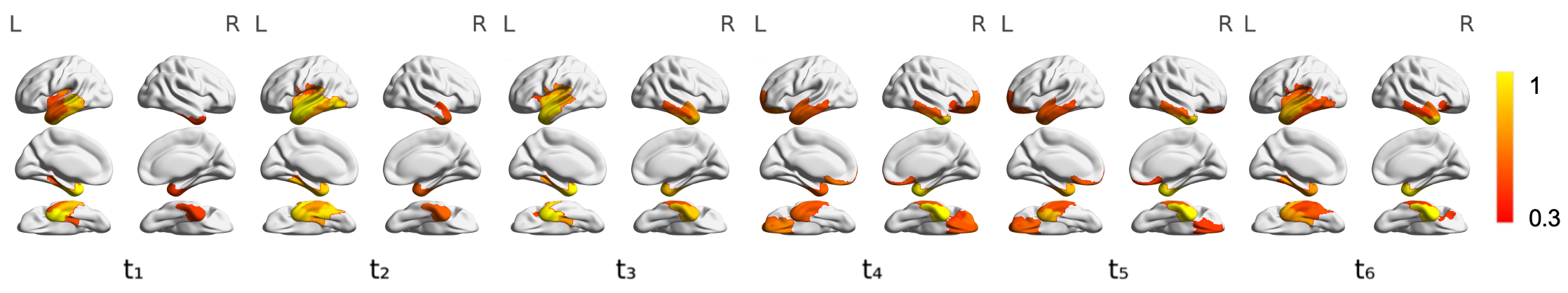}
    \vspace{-20pt}
    \caption{Longitudinal visualization of salient ROIs in the progressive group across six follow-up visits. Color indicates regional importance (yellow: higher; red: lower) among the top-20 ROIs at each time point.}
    \label{fig:visual}
\end{figure*}

\begin{figure}[h!]
    \centering
    \includegraphics[width=\columnwidth]{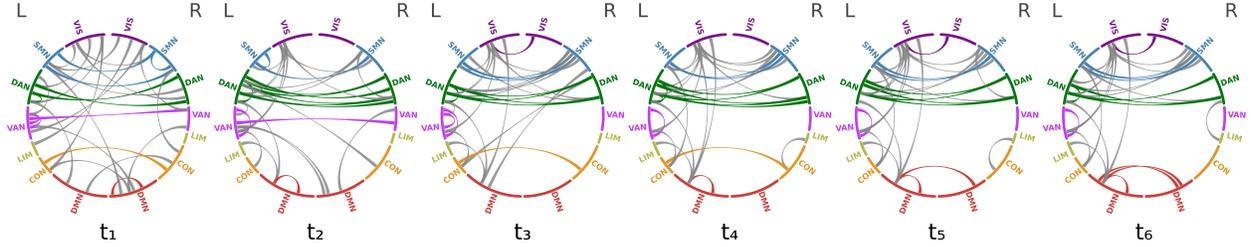}
     \vspace{-18pt}
    \caption{Top-30 discriminative functional connections between stable and progressive groups across six longitudinal time points ($t_1$--$t_6$), identified via FDR-corrected two-sample $t$-tests ($p < 0.05$) on learned hyperedge importance probabilities. ROIs are grouped into seven large-scale neural systems based on the Schaefer100 (7-network) parcellation: visual (VIS), somatomotor (SMN), dorsal attention (DAN), ventral attention (VAN), limbic (LIM), frontoparietal control (CON), and default mode network (DMN). The intra-network connections are colored by their respective network and inter-network connections are colored in grey.}
    \label{fig:chord}
\end{figure}

\textbf{\textit{Interpretation Analysis.}} To examine the clinical relevance of the learned representations, we visualize the top-20 salient ROIs for progressive subjects across six follow-up visits on the OASIS-3 dataset (Fig.~\ref{fig:visual}). The visualization reveals a clear temporal evolution of important regions. Early visits consistently highlight the parahippocampal cortex, a well-established early biomarker of AD~\cite{echavarri2011atrophy,van2000parahippocampal}. At later stages, salient regions expand toward prefrontal and temporo-parietal cortices, consistent with the posterior-to-anterior spread of tau pathology described in Braak staging~\cite{braak2006staging,vogel2021four}. These findings indicate that SDE-HGNN captures temporally evolving functional reorganization associated with AD progression rather than producing static importance patterns.

To further analyze connectivity changes, we examine the learned hyperedge importance probabilities and perform two-sample $t$-tests to identify statistically significant connectivity differences between stable and progressive groups on OASIS-3. Fig.~\ref{fig:chord} visualizes top-30 discriminative connections across six longitudinal time points. ROIs are grouped into seven large-scale functional networks based on the Schaefer100 (7-network) parcellation: visual (VIS), somatomotor (SMN), dorsal attention (DAN), ventral attention (VAN), limbic (LIM), frontoparietal control (CON), and default mode network (DMN). Across all visits, the most discriminative connections primarily involve the DAN, DMN, and SMN. Notably, interactions between DAN and DMN appear even at baseline, suggesting that early disruption of these opposing systems may serve as an early functional marker of AD. As disease progresses, connectivity differences increasingly involve prefrontal and temporo-parietal regions, consistent with known disease progression patterns.

\begin{figure}[t]
    \centering
    \vspace{-10pt}
    \includegraphics[width=0.9\columnwidth]{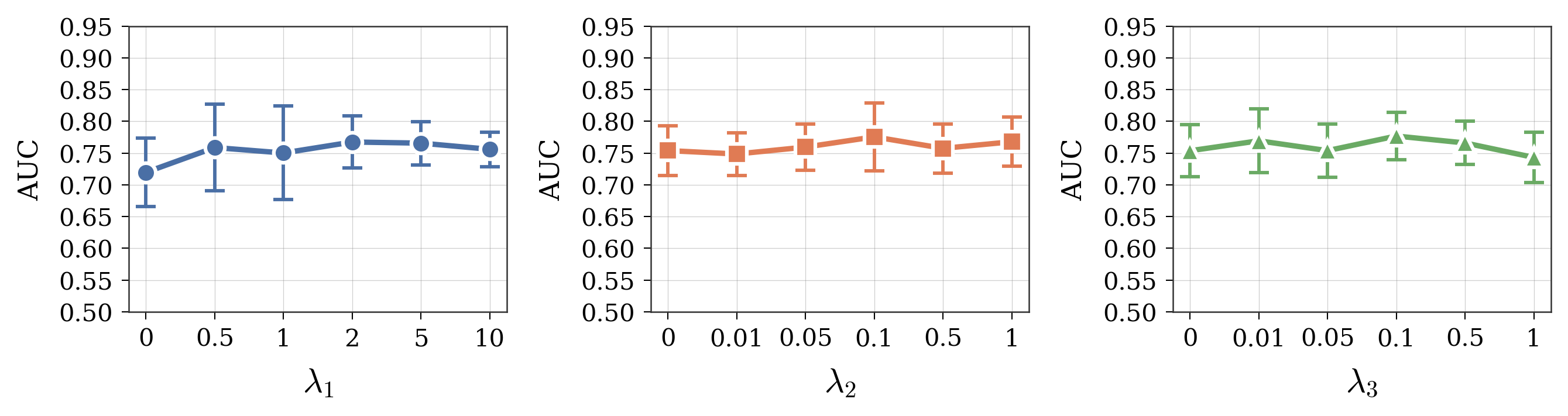}
    \vspace{-10pt}
    \caption{Sensitivity analysis of three loss weighting hyperparameters.}
    \label{fig:sensitivity}
\end{figure}

\textbf{\textit{Hyperparameter Sensitivity Analysis.}}
We analyze the sensitivity of model performance to the three loss weighting parameters in Eq.~(\ref{eq:loss}): $\lambda_1$ (mutual information loss), $\lambda_2$ ($\ell_1$ sparsity regularization), $\lambda_3$ (entropy regularization). For each parameter, we vary its value across six levels while fixing the remaining parameters at their optimal settings. Fig.~\ref{fig:sensitivity} reports the resulting AUC values. Across all three parameters, we observe an inverted-U trend, where performance improves as the parameter increases from a small value, peaks at an intermediate value, and gradually decreases for larger values. The best performance is obtained at: $\lambda_1 = 2$, $\lambda_2 = 0.1$, and $\lambda_3 = 0.1$. These results indicate that moderate regularization provides the best balance between feature selection and predictive performance.

\textbf{\textit{Computational Efficiency.}} We evaluated the computational cost of the three strongest baseline methods considered in this study—WHGCN, SDEGCN, and the proposed SDE-HGNN—on the OASIS-3 dataset with six longitudinal time points using five-fold cross-validation on NVIDIA RTX A6000 GPUs. The end-to-end runtime was 5.9 minutes for WHGCN, including 5.8 minutes for hyperedge-weight precomputation, compared with 75.7 minutes for SDE-HGNN (9.08 s/epoch) and 99.6 minutes for SDEGCN (11.95 s/epoch). While WHGCN achieves the shortest runtime due to its simpler discrete-time formulation, SDE-HGNN consistently delivers superior predictive performance while requiring less computation than SDEGCN. These results demonstrate that SDE-HGNN achieves a favorable trade-off between predictive accuracy and computational efficiency.

\section*{Conclusion}
\label{sec_conclusion}
In this work, we proposed SDE-HGNN, an stochastic differential equations-driven spatio-temporal hypergraph neural network framework for modeling irregular longitudinal fMRI connectomes in Alzheimer's disease. By integrating continuous-time temporal modeling with higher-order hypergraph representations, SDE-HGNN captures both disease progression dynamics and coordinated multi-region functional interactions under irregular longitudinal sampling. Experiments on the OASIS-3 and ADNI cohorts demonstrate consistent improvements over state-of-the-art graph and hypergraph baselines in both progression prediction and diagnostic classification tasks. In addition, the proposed sparsity-based importance learning mechanism identifies salient brain regions and discriminative connectivity patterns associated with AD progression, providing interpretable insights into disease-related brain network alterations. Beyond predictive performance, SDE-HGNN may serve as a promising framework for longitudinal clinical decision support by providing progression-risk estimates together with interpretable neuroimaging biomarkers. Future work will evaluate the framework in larger longitudinal cohorts and investigate its integration with multimodal biomarkers.

\section*{Acknowledgments}
This work was supported in part by NIH (R01LM013519, RF1AG077820), NSF (IIS-2319451, MRI-2215789), DOE (DE-SC0025801), and Lehigh University (CORE and RIG).

\makeatletter
\renewcommand{\@biblabel}[1]{\hfill #1.}
\makeatother

\bibliographystyle{vancouver}
\bibliography{ref_ry}
\end{document}